\documentclass[
twocolumn
]{ceurart}

\usepackage{listings}
\begin{document}



\title{Column Type Annotation using ChatGPT}


\author[1]{Keti Korini}[%
orcid=0000-0002-2158-0070,
email=kkorini@uni-mannheim.de,
]
\cormark[1]

\author[1]{Christian Bizer}[%
orcid=0000-0003-2367-0237,
email=christian.bizer@uni-mannheim.de,
]
\address[1]{Data and Web Science Group, University of Mannheim, Mannheim, Germany}

\cortext[1]{Corresponding author.}


\begin{abstract}
  Column type annotation is the task of annotating the columns of a relational table with the semantic type of the values contained in each column. Column type annotation is an important pre-processing step for data search and data integration in the context of data lakes. State-of-the-art column type annotation methods either rely on matching table columns to properties of a knowledge graph or fine-tune pre-trained language models such as BERT for column type annotation. In this work, we take a different approach and explore using ChatGPT for column type annotation. We evaluate different prompt designs in zero- and few-shot settings and experiment with providing task definitions and detailed instructions to the model. We further implement a two-step table annotation pipeline which first determines the class of the entities described in the table and depending on this class asks ChatGPT to annotate columns using only the relevant subset of the overall vocabulary. Using instructions as well as the two-step pipeline, ChatGPT reaches F1 scores of over 85\% in zero- and one-shot setups. To reach a similar F1 score a RoBERTa model needs to be fine-tuned with 356 examples. This comparison shows that ChatGPT is able deliver competitive results for the column type annotation task given no or only a minimal amount of task-specific demonstrations.
  \end{abstract}

\begin{keywords}
  Table Annotation \sep
  Column Type Annotation \sep
  ChatGPT \sep
  Large Language Models \sep
  Prompt Engineering
\end{keywords}

\maketitle

\section{Introduction}
\begin{figure}
  \centering
  \includegraphics[width=\linewidth]{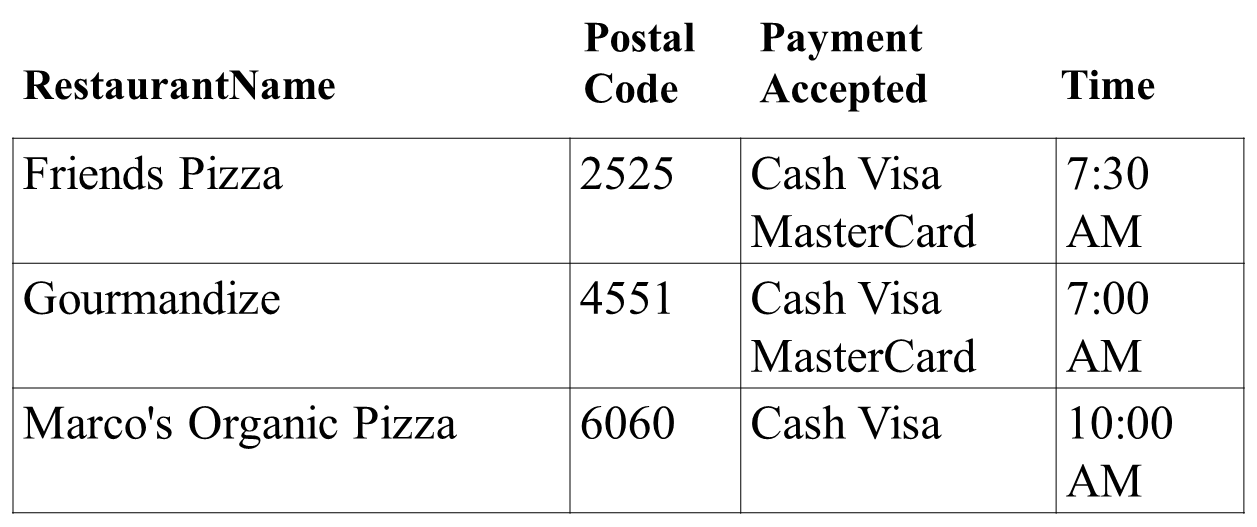}
  \caption{Example table describing restaurants. The semantic types that are assigned to each column by CTA are shown in bold above the columns.}
  \label{fig:cta}
\end{figure}

Table annotation refers to the task of discovering semantic information about elements of a table such as columns, relationship between columns, and entities contained in table cells~\cite{taSurvey2023}\footnote{\url{https://paperswithcode.com/task/table-annotation}}.  The task of \textit{Column Type Annotation} (CTA) is a sub-task of table annotation which focuses on annotating the columns of a relational table with the semantic type of the values contained in each column given a predefined set of semantic types. CTA is an important pre-processing step for data search~\cite{chapman2020dataset}, knowledge base completion~\cite{ritze-profiling-2016}, and data integration in the context of data lakes~\cite{datalakesSurvey}. Figure \ref{fig:cta} shows a table describing restaurants. A CTA method would examine the cell content and for instance conclude that the first column should be annotated with the semantic type ``RestaurantName", while the third column containing payment methods would be labeled as ``PaymentAccepted".

A wide range of CTA methods has been proposed in the last years~\cite{taSurvey2023} : One line of work relies on linking the entities in a table to a knowledge graph (KG) and determines the column types based on the types of the linked entities afterwards~\cite{jimenez-ruiz_semtab_2020}. A second line of work relies on pre-trained language models (PLM) such as BERT \cite{devlinBERTPretrainingDeep2019} or RoBERTa~\cite{liu_roberta_2019}. The models are either directly fine-tuned for the CTA task~\cite{suhara2022annotating} or are further pre-trained on tabular data and fine-tuned for the CTA task afterwards~\cite{deng_turl_2022, iida-etal-2021-tabbie}. In order to reach good performance, most state-of-the-art CTA methods require significant amounts of task-specific training data.
Large language models (LLMs)~\cite{zhao2023survey} such as GPT~\cite{brown2020language}, ChatGPT~\cite{ouyang2022training}, PaLM~\cite{chowdhery2022palm}, or BLOOM~\cite{yong2022bloom+} have the potential to reduce the required amount of task-specific training data, or make task-specific training data even completely obsolete. Due to being pre-trained on huge amounts of text as well as due to emergent effects resulting from the model size~\cite{wei2022emergent}, LLMs often have a better zero- and few-shot performance compared to PLMs such as BERT and are also more robust concerning unseen examples~\cite{brown2020language}. 

Initial research on exploring the potential of LLMs for data integration tasks, such as schema matching, entity matching, data imputation, and value normalization was conducted by Narayan et al.~\cite{foundationalWrangleVLDB2022} and Jaimovitch-Lopez et al. \cite{jaimovitch-lopez_can_2022}.
To the best of your knowledge, using LLMs for table annotation has not been explored yet. This paper fills this gap with an initial, explorative study on using ChatGPT~\cite{ouyang2022training} for column type annotation. 
More specifically, the contributions of this paper are the following:

\begin{enumerate}
    \item We experiment with different prompt designs for column type annotation using a subset of the SOTAB benchmark~\cite{korini_sotab_2022}. 
    \item We investigate the impact of providing step-by-step instructions, using message roles, as well as in-context learning on the performance of ChatGPT for the CTA task.
    \item We propose a two-step annotation pipeline which enables ChatGPT to deal with large semantic type sets.
    \item We compare the performance of ChatGPT to the performance of RoBERTa~\cite{liu_roberta_2019} and DODUO~\cite{suhara2022annotating}, a state-of-the-art CTA method, using different amounts of training data.
\end{enumerate}

The paper is organized as follows: Section \ref{sec:exp-set} describes the experimental setup. Section \ref{sec:prompt} sets a baseline by employing three simple prompts in a zero-shot setup. In Section \ref{sec:instr}, we experiment with providing explicit instructions on how to perform CTA to the model; in Section \ref{sec:chat}, we experiment with using ChatGPT's message roles; in Section \ref{sec:in-context}, we switch to a few-shot setup and explore in-context learning for the CTA task. In Section \ref{sec:two-step}, we experiment with a two-step pipeline in order to cover larger vocabularies. Section \ref{sec:baselines} compares our zero- and few-shot results for ChatGPT to the results of a RoBERTa model and DODUO. Section \ref{sec:rel-work} discusses related work.

All data and code used in this paper are available at the project github\footnote{\url{https://github.com/wbsg-uni-mannheim/TabAnnGPT}} meaning that all experiments can be replicated.

\section{Experimental Setup}
\label{sec:exp-set}
This section describes the dataset and the language model that we use for the experiments and explains how we calculate F1 based on the model's answers.

\textbf{Dataset.} We use the SOTAB benchmark~\cite{korini_sotab_2022} for our experiments. The SOTAB benchmark consists of tables that have been extracted from different websites and are annotated using terms from the schema.org vocabulary\footnote{\url{https://schema.org/}}. The full test set of the benchmark consists of 15,040 columns and the full training set 130,471 columns which are annotated using 91 different semantic types. For our explorative study, we down sample SOTAB in order to keep the cost of using ChatGPT via the OpenAI API in an acceptable range\footnote{\$0.002 per 1000 tokens. See \url{https://openai.com/pricing}}. For building our training and test sets we select tables from the original training and test sets. The selected tables belong to four different domains: music, hotels, events, and restaurants. We also down sample the label space to consist of 32 semantic types. 
Overall, we select 62 tables for our training set containing 356 columns which are labeled with their semantic types and 41 tables for the test set containing 250 labeled columns. We manually verify the annotations for all chosen tables. The columns contain three different types of values: textual, date and numerical values, with textual being the most frequent type. Table \ref{tab:stats} provides statistics comparing the complete SOTAB benchmark datasets and our down-sampled subsets. Table \ref{tab:types} lists the semantic types that we use in the experiments. Note that we require the models to be able to distinguish different types of names, e.g. MusicRecordingName, RestaurantName, HotelName, and EventName, as well as closely related text columns such as entity description and entity review.  

\begin{table}
  \caption{Statistics of the SOTAB benchmark and the down-sampled datasets.}
  \label{tab:stats}
  \begin{tabular}{lcccc}
    \toprule
    &Set&Tables&Columns&Labels\\
    \midrule
    \multirow{2}{*}{\makecell[l]{SOTAB CTA \\ complete}} & Training & 46,790 & 130,471 & 91 \\
    & Test & 7,026 & 15,040 & 91 \\
    \multirow{2}{*}{\makecell[l]{Down-sampled \\ datasets}}& Training & 62 & 356 & 32 \\
    & Test & 41 & 250 & 32 \\

  \bottomrule
\end{tabular}
\end{table}

\begin{table}
  \caption{Overview of the semantic types that are used for table annotation in the experiments grouped by domain.}
  \label{tab:types}
  \begin{tabular}{ll}
    \toprule
    Domain&Labels\\
    \midrule
    Music Recording &  \makecell[l]{MusicRecordingName, Duration, \\ ArtistName, AlbumName} \\
    \hline    
    Restaurants & \makecell[l]{RestaurantName,  PriceRange, \\ 
    AddressRegion, Country, Telephone, \\
    PaymentAccepted, PostalCode,\\ Coordinate, DayOfWeek, Time, \\ RestaurantDescription, Review } \\
    \hline
    Hotels & \makecell[l]{HotelName, PriceRange,  Telephone, \\ FaxNumber,  Country, Time,\\  PostalCode,  AddressLocality,  email,\\ LocationFeatureSpecification, \\ HotelDescription,  Review, Rating,\\ PaymentAccepted, Photograph} \\
    \hline
    Events &  \makecell[l]{EventName, Date, DateTime, \\ EventStatusType, EventDescription,\\ EventAttendanceModeEnumeration,\\ Organization, Currency, Telephone}\\
  \bottomrule
\end{tabular}
\end{table}

\textbf{Language Model.} We use ChatGPT version ``gpt-3.5-turbo-0301" for our experiments. We use the \textit{Langchain} python package\footnote{https://python.langchain.com/en/latest/} to access the model via the OpenAI API and set the temperature parameter to 0 in order to lower the variability of the answers given the same input.

\textbf{Evaluation.} We employ a multi-class problem setup for column type annotation, meaning that each column can be annotated with exactly one label. The metrics used for the evaluation are precision, recall and micro-F1 score. We use the micro-F1 score that is less influenced by the different number of examples for each label and their individual performances. The model sometimes answers using not exactly the requested terms but synonyms of the requested terms. We manually collect such synonyms from several test runs into a dictionary and count answers that are contained in this dictionary as correct in the evaluation. Altogether, the dictionary contains 27 synonyms for the 32 labels. Example of synonyms are: ``Check-in Time" which can be mapped to the label ``Time" or ``Amenities" which can be mapped to the label ``LocationFeatureSpecification".


\section{Simple Prompts}
\label{sec:prompt}

Designing good prompts is the key challenge for successfully using LLMs for prediction tasks as the choice of formulations~\cite{zhao2021calibrate} and even the choice of specific words~\cite{webson-pavlick-2022-prompt} can strongly affect model performance.
In order to establish baselines, we evaluate three different approaches to formulate simple prompts for the CTA task. The first two prompts ask ChatGPT to determine the semantic type of single columns contained in the table. The third prompt instructs ChatGPT to determine the types of all columns of the table at once. All three prompts start with a guiding sentence that instructs the model to answer according to the task given and in case that
it does not know the answer, it should reply with "I don't know".

\begin{figure}
  \centering
  \includegraphics[width=\linewidth]{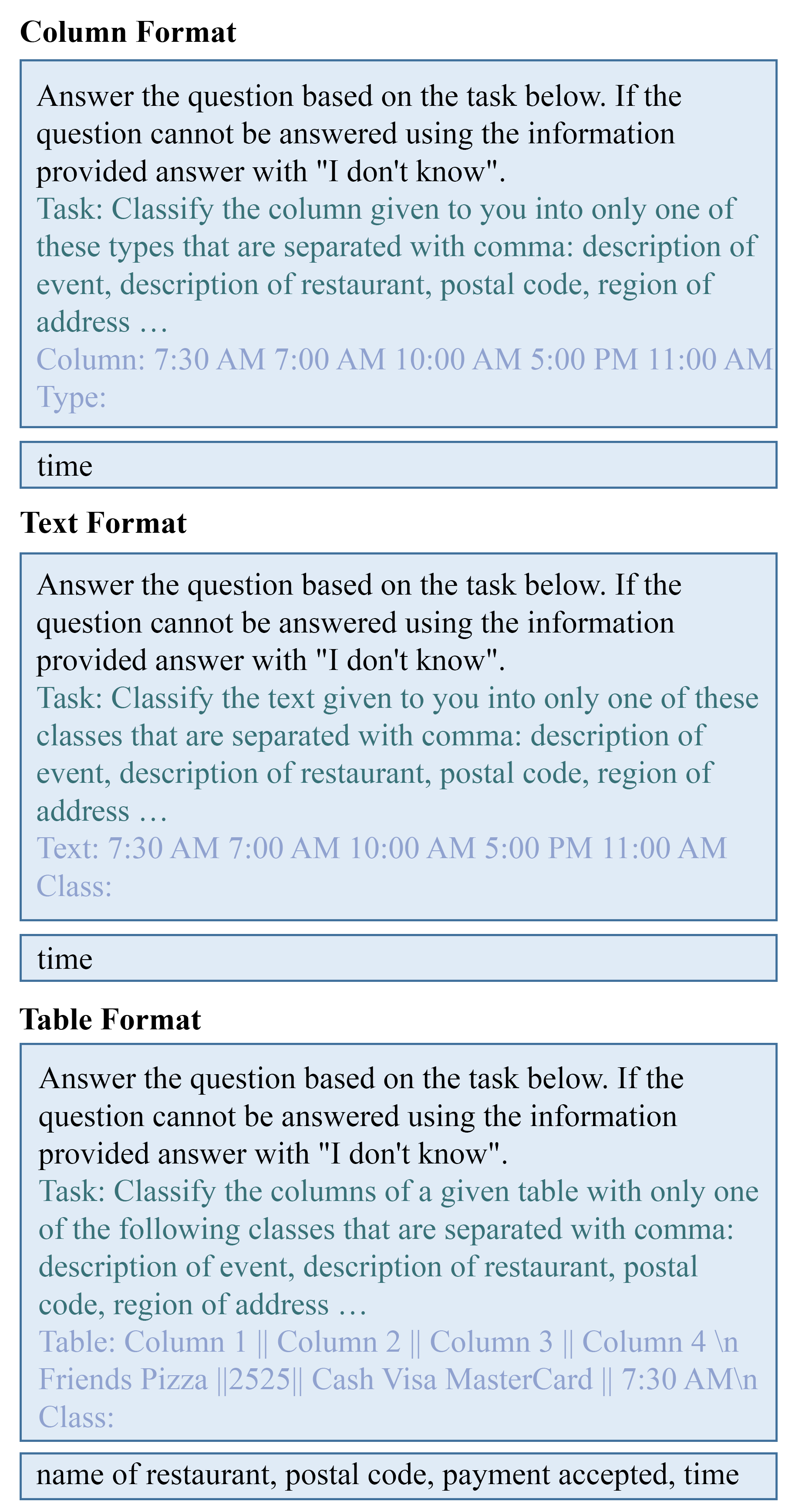}
  \caption{Prompt examples for column, text, and table format}
  \label{fig:formats}
\end{figure}

\textbf{Column.} The column prompt uses terminology directly related to the CTA task. The prompt starts with a task description ``\emph{Classify the column given to you into one of these types which are seperated by comma}". The task description is followed by the list of all the 32 types in the labels set. Afterwards, the column that should be annotated is included into the prompt. The column is represented by the keyword ``\emph{Column:}" followed by the concatenation of the column values in the first five rows of a table. We use the word ``\emph{Type:}" to instruct the model to predict the semantic type of the column. For each test example, the prediction of the model in this case is either a single word belonging to a label in the label set, a word not contained in the label set or ``I don't know". An example of a prompt using the described format is found in the upper part of Figure \ref{fig:formats}. The example shows how the fourth column in the table in Figure \ref{fig:cta} is passed to ChatGPT. The blue box below the prompt contains an example of an answer by ChatGPT, in this case the correct semantic type \emph{time}.

\textbf{Text.} In order to test whether ChatGPT performs better if the CTA task is presented as a generic text classification task, the second prompt uses generic  terms related to text classification. We formulate the task description part of the prompt as ``\emph{Classify the text given to you into one of these classes that are separated with comma}", again followed by the list of all semantic types. The test example is again represented as the concatenation of the column values of the first five rows. In this prompt, we use the word ``\emph{Class:}" to instruct the model to return one of the classes. An example of a prompt using the text format is found in the middle of Figure \ref{fig:formats}.

\textbf{Table.} In addition to examining the content of the column to be annotated itself, it is often also necessary to consider the content of the other columns in a table in order to assign the correct semantic type to a column. In order to allow ChatGPT to exploit the context of a column for its decisions as well as allowing the model to consider dependencies between  annotations, we include complete tables, inputted row by row, into the prompt and ask ChatGPT to annotate all columns of the table at once. We formulate the task as ``\emph{Classify the columns of a given table with one of the following classes}". As ChatGPT version ``gpt-3.5-turbo-0301" has a token limit of 4097 tokens, we select only the five first rows of a table and turn the table into string format as follows: we separate different cells with the notation ``||" and we divide different rows with the notation ``\textbackslash n" (e.g. Table \ref{fig:cta} would be \emph{Column 1 || Column 2 || Column 3 || Column 4 || \textbackslash nFriends Pizza || 2525 || Cash Visa MasterCard || 7:30 AM ||\textbackslash n...}). The prediction of the model in this case returns a string separated with commas which contains in the order of the columns the type prediction for all columns in the input. In rare cases that the model replies with full sentences, the label would be contained in quotation marks, so we extract the text within the quotation marks and check if the answer can be linked to our label set using a dictionary. An example of a prompt using the table format is found in the lower part of Figure \ref{fig:formats}.

\textbf{Experimental Results.} The results of querying ChatGPT for all 250 columns of the test set using the different prompt formats can be found in the upper section of Table \ref{tab:format}. Both text and column formats achieve a similar performance of 45-47\% micro-F1 score, with text classification performing approx. 1\% better. The table format scores roughly 8\% less than the column format. This indicates that ChatGPT was partly confused by the longer input and the more complex task of annotating multiple columns at once.

\begin{table}
  \caption{Results for the three different prompt format in the zero-shot setting: text, column and table. ``+inst" indicates the experiments where instructions were added, while ``roles" indicates the experiments in which the message roles were used. We report precision (P), recall (R), and micro-F1 (F1) . The $\Delta$ F1 shows the difference between the micro-F1 score of our baseline model (simple column format) to each experiment.}
  \label{tab:format}
  \begin{tabular}{lccccl}
    \toprule
    Format&P&R&F1&$\Delta$ F1\\ 
    \midrule
    column & 47.70 & 31.25 & 45.85& - \\ 
    text & 46.38 & 33.97 & 47.02& +1.17\\ 
    table & 41.08 & 32.38 & 37.90& -7.95 \\ 
    \hline
    column+inst& 72.00 & 51.18 & 62.27& +16.42  \\ 
    text+inst & 63.94 & 47.20 & 57.95& +12.10 \\ 
    table+inst & 81.88 & 76.79 & 80.16& +34.31  \\ 
    \hline
    column+inst+roles & 86.99 & 69.95 & 78.61& +32.76 \\ 
    text+inst+roles & 83.68 & 67.13 & 74.15& +28.30 \\ 
    table+inst+roles & 85.91 & 82.01 & 85.25& +39.40 \\ 
  \bottomrule
\end{tabular}
\end{table}
 
\section{Providing Explicit Instructions}
\label{sec:instr}

\begin{figure}
  \centering
  \includegraphics[width=\linewidth]{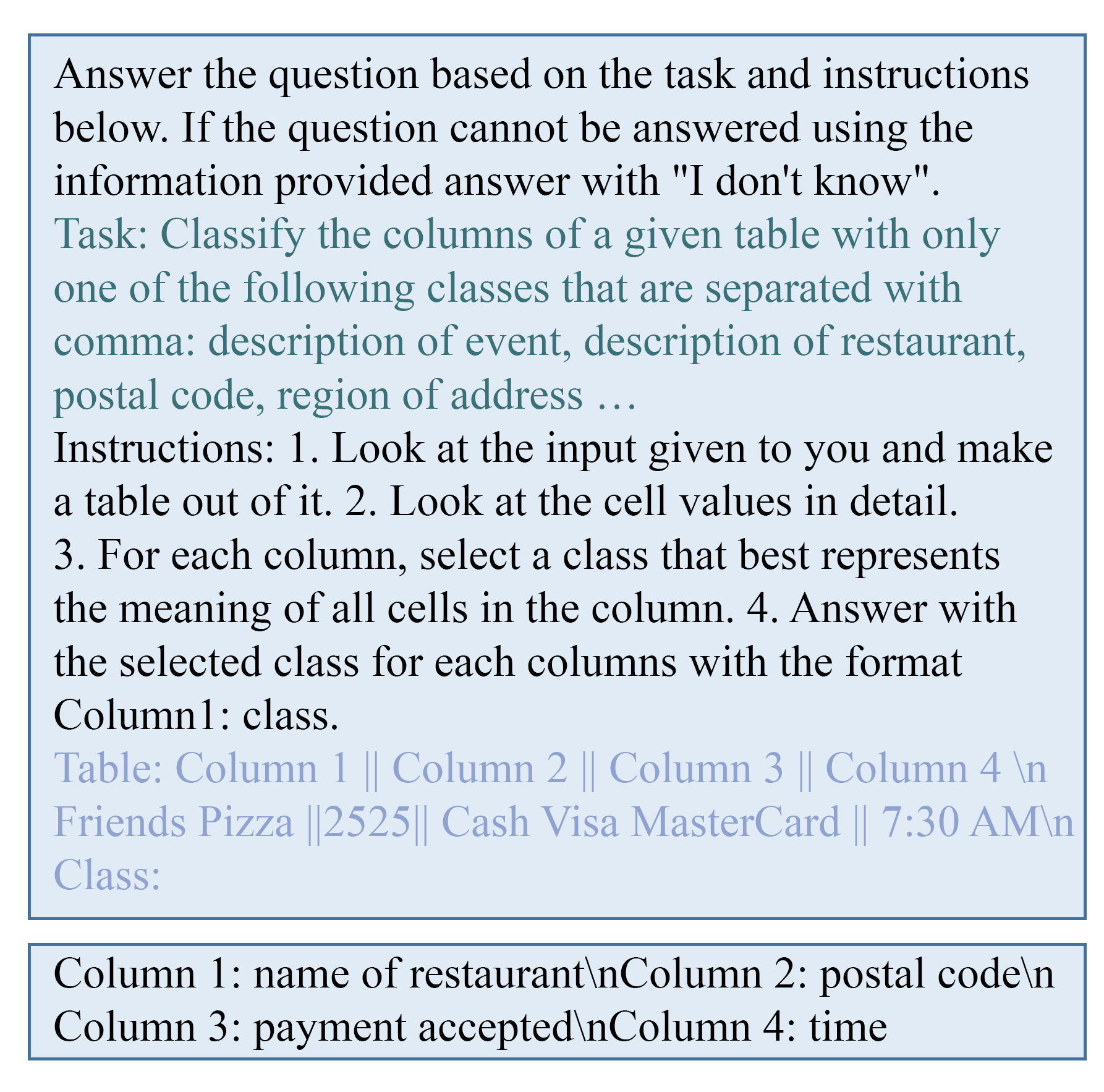}
  \caption{Instructions for the table format.}
  \label{fig:instr}
\end{figure}

Previous work has shown that supporting the model via the prompt in decomposing a task into several steps can improve model performance~\cite{wei_chain--thought_2022}. Inspired by this work, we experiment with providing step-by-step instructions to ChatGPT on how to approach the CTA task:
We ask the model to first analyze the input is it given, afterwards it should select the class/type that best represents the meaning of the input, and should then reply with the corresponding class/type. We modify our original prompt template by adding an instruction part after the task definition. For the table format, an example of an extended prompt is shown in Figure \ref{fig:instr}, while for the column and text formats we list the instructions below:

\begin{itemize}
    \item \textbf{column:} 1. Look at the column and the types given to you. 2. Examine the values of the column. 3. Select a type that best represents the meaning of the column. 4. Answer with the selected type.
    \item \textbf{text:} 1. Look at the text and the classes given to you. 2. Examine the values of the text. 3. Select a class that best represents the meaning of the text. 4. Answer with the selected class.
\end{itemize}

One of the important parts of these instructions in regards to the table format is that we instruct the model to generate a table out of the input that it has been given before proceeding with classification. This instruction was added with the motivation to make the model understand that it is working with a table and building the table out of the input would give a better understanding of the rows and columns that the table is made of. The model is also instructed in the last point about the desired reply format in order to ease the parsing of the model's answers.

\textbf{Experimental Results.} The results of the experiments including explicit instructions are provided in the second part of Table \ref{tab:format}  (indicated by the ``+inst" notation). We notice that by providing instructions to the model, the performance increases by between 12 to 35\% in micro-F1 score. The result that is impacted most by the instructions is the result for the table format which jumps to an F1 score of 80\%. The instruction to turn the input into a table seems to help ChatGPT a lot in understanding the table content. We also observe that given the instructions, the multi-column table annotation approach clearly outperforms the two single column approaches by 18 and 22\% F1. 

\section{Using Message Roles}
\label{sec:chat}

\begin{figure}
  \centering
  \includegraphics[width=\linewidth]{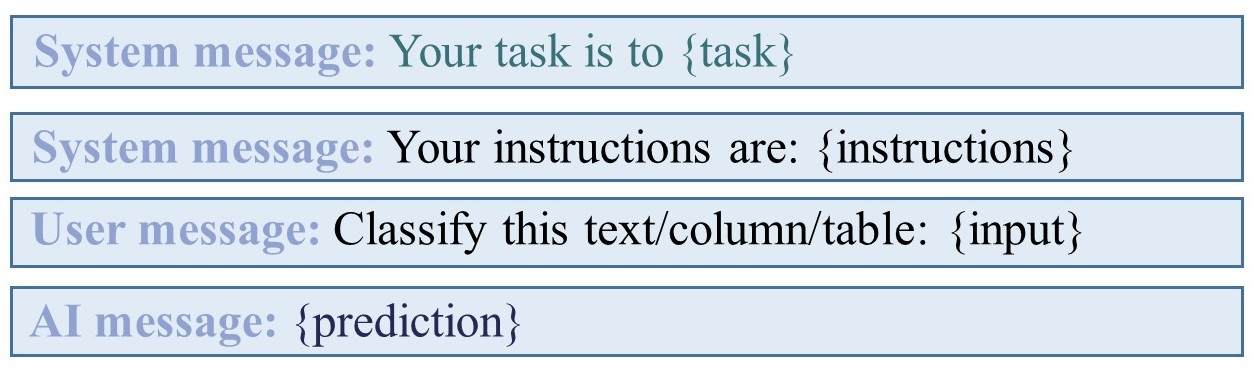}
  \caption{Message templates for the three formats.}
  \label{fig:chat}
\end{figure}

Chat models such as \emph{gpt-3.5-turbo} and \emph{gpt-4} offer message roles to distinguish between \emph{System}, \emph{User}, and \emph{AI} messages in a conversation. System messages are used to set the general behavior of the model; user messages are used by the user to pass a query or a task to the model, and AI messages contain the responses of the model\footnote{https://platform.openai.com/docs/guides/chat}. The previous experiments did not use message roles. In this section we test whether using  message roles improves the performance for the CTA task. As illustrated in Figure \ref{fig:chat}, we use system messages to pass task descriptions (see Section \ref{sec:prompt}) and instructions (see Section \ref{sec:instr}) to the model. We use a user message to pass the actual annotation task to the model, which answers with an AI message.

\textbf{Experimental Results.} The results of running the experiments using the three base prompts from Section \ref{sec:prompt} together with the instructions from Section \ref{sec:instr} and the roles described above are presented in the lower part of Table \ref{tab:format} (indicated with the word ``roles"). From the results, we can see an increase of 28\% to 39\% in micro-F1 score compared to the column format baseline and an increase of 5\% to 16\% compared to the results of the instruction prompts presented in the middle section of the table. Thus, using the message roles offered by \emph{gpt-3.5-turbo} proved beneficial in all cases and we thus also use them in the following experiments.

\begin{figure}
  \centering
  \includegraphics[width=\linewidth]{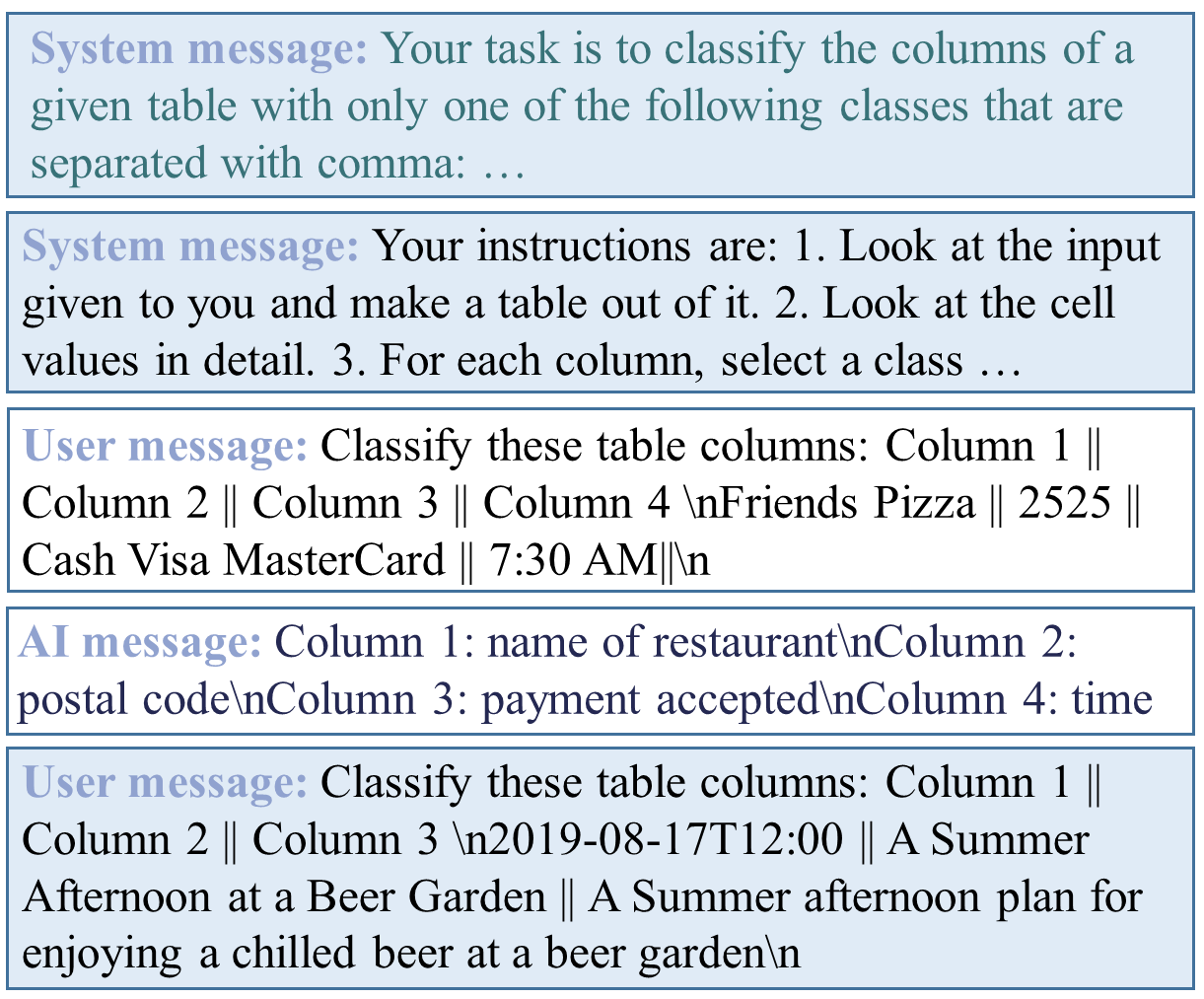}
  \caption{Example of one-shot table format messages. The demonstration is shown in the white boxes using a user and AI message.}
  \label{fig:incontext}
\end{figure}

\section{In-Context Learning}
\label{sec:in-context}

The performance of LLMs can be improved by providing them demonstrations of the task that they are supposed to perform as part of the prompt~\cite{brown2020language}.  We continue our experiments by providing  task demonstrations to the model. Showing demonstrations (or training examples) to the model is also known as few-shot learning, where shots are the number of demonstrations shown. All previous experiments were zero-shot experiments as no demonstrations are shown to the model. We experiment with a one-shot and a five-shot setup. A question that arises in this case is how to choose which examples to show to the model. Since in our setup CTA is a multi-class classification problem, the training set is composed of multiple examples per label. However, we can not pick demonstrations by relevancy (e.g. show an example of restaurant name for predicting a column about restaurant names) as this would leak information about the ground truth labels. For this reason, we decide to pick demonstrations randomly from the training set, without considering the class of the entities described in the table. In the case of column and text format, the demonstrations follow the format of the test example and therefore are columns represented by concatenating the values of the first five rows of a table. For the table format an example is a randomly chosen table containing the first five rows of the original table.

As shown in Figure \ref{fig:incontext}, we use a user message to present the demonstration task to the model (first user message in Figure \ref{fig:incontext}) and an AI message to show the model the expected answer, e.g. the ground truth labels represented using the expected format (first AI message in Figure \ref{fig:incontext}). Afterwards, we present the actual test example using a further user message (last user message in Figure \ref{fig:incontext}). 

\begin{table}
  \caption{Average results over three runs for the three format types by providing demonstrations. ``shots=5" means that five demonstration are provided. The $\Delta$ F1 shows the difference between the micro-F1 score of our baseline model (column format zero-shot shown in the first line) to each experiment.}
  \label{tab:incontext}
  \begin{tabular}{lccccl}
    \toprule
    Format&shots&P&R&F1&$\Delta$ F1\\
    \midrule
    column & 0 & 47.70 & 31.25 & 45.85& - \\
    \hline
    column & 1 & 88.70 & 82.02 & 84.57 & +38.72\\
    column & 5 & 90.15 & 86.03 & 88.49 & +42.64\\
    text & 1 & 81.96 & 71.89 & 75.16 & +29.31 \\
    text & 5 & 88.32 & 81.46 & 84.24 & +38.29\\
    table & 1 & 88.67 & 84.81 & 88.44 & +42.59 \\
    table & 5 & 87.51 & 85.28 & 88.83 & +42.98\\
  \bottomrule
\end{tabular}
\end{table}

\textbf{Experimental Results.} Table \ref{tab:incontext} presents the results of running the experiments with prompts containing the instructions, using message roles, and containing either one or 5 demonstrations (1-shot or 5-shot). The reported scores are averages of three runs as the demonstrations are randomly picked at runtime. For all three formats the inclusion of examples improves the performance of the model by 29\% to 42\% compared to the zero-shot column format baseline. Compared to the experiments using instructions and roles (see lower part of Table \ref{tab:format}), providing demonstrations increases the performance by a further 1-10\% F1 score. Generally, we notice that with the increase of the number of shots (demonstrations) the F1 score also improves with the exception of the table format where we observe only a slight 0.39\% increase in the 5-shot case, which might result from the model being confused by the length of the prompt including 5 tables. The highest increase was observed with the column and text format in the 5-shot case where the performance increases by 10\%. Experiments with more than five-shots were not conducted as the token limit of 4097 tokens was usually surpassed when showing more than 5 table demonstrations in the case of the table format. In the zero-shot table format setting the average token length of the prompt used for annotating one test example is 550 tokens (without including the response from the model). This increases to an average of 900 in the one-shot setting and up to an average of 2320 when 5 table demonstrations are given.

\textbf{Out-of-Vocabulary Answers.} 
ChatGPT sometimes ignores the instruction to use terms from the label space but answers using different terms. For the zero-shot prompts, on average 27 out of the 250 answers did not exactly match the label space. Using the dictionary of synonyms (see Section \ref{sec:exp-set}), on average 4 of these answers could be mapped to one of the labels. The amount of out-of-vocabulary answers decreases to an average of 12 out of 250 in the few-shot setting, where 6 of them could be mapped to the label space. 

\section{Two-Step Pipeline}
\label{sec:two-step}

\begin{figure}
  \centering
  \includegraphics[width=\linewidth]{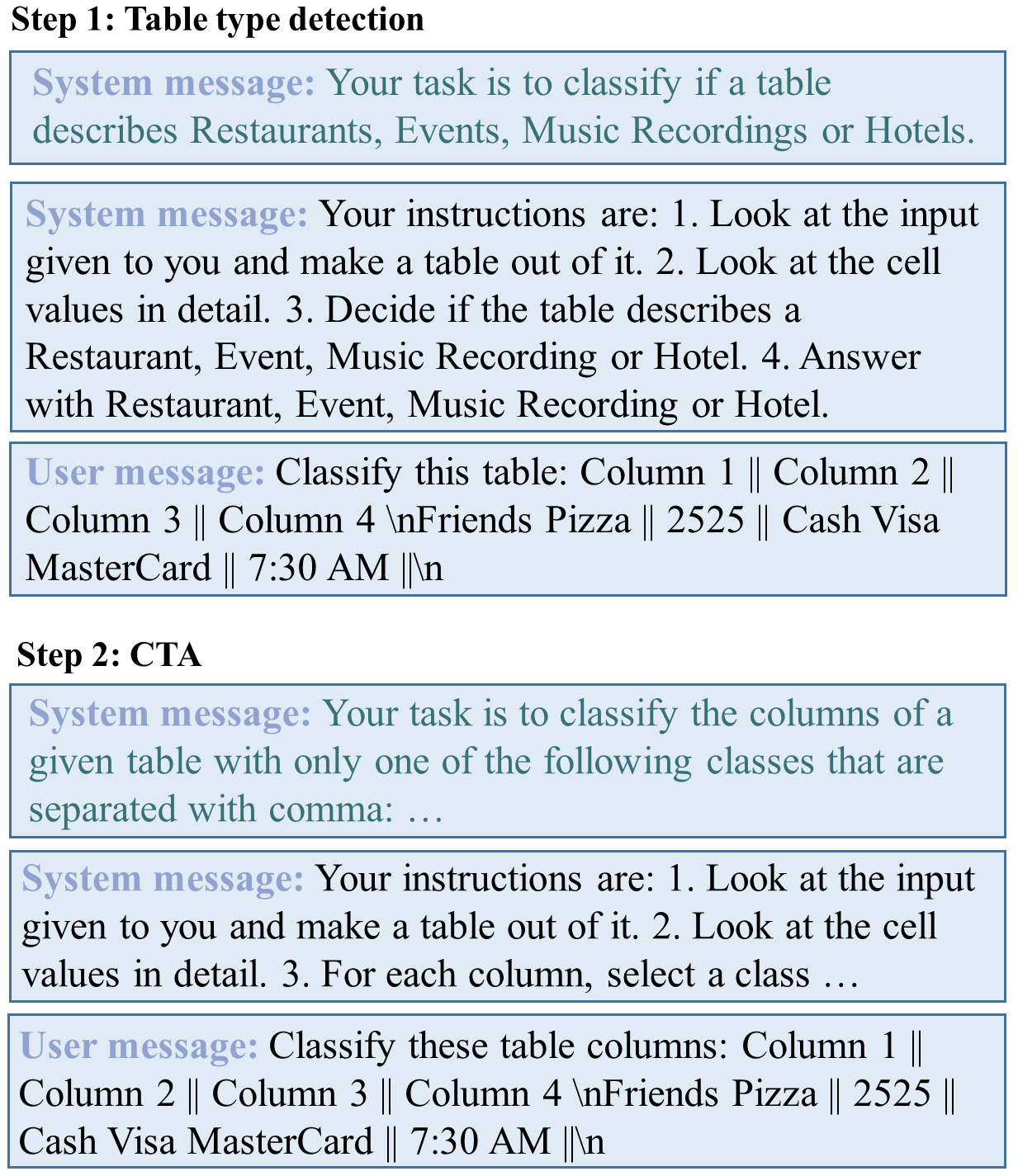}
  \caption{Example of zero-shot setup for the two-step pipeline.}
  \label{fig:twostep}
\end{figure}

The task description of all prompts that we presented so far contains the complete list of the semantic types that should be used for the annotation (32 types in our case). Other annotation use cases might involve larger label spaces. For instance, the complete labels space of the SOTAB CTA benchmark~\cite{korini_sotab_2022} consists of 91 semantic types (see Table \ref{tab:stats}); the label space of the WikiTables dataset~\cite{deng_turl_2022} consists of 255 types. In order to prevent needing to add the complete list of semantic types to the prompts and therefore allowing the prompts to be used with larger label spaces, we propose a two-step pipeline which exploits schema information about the semantic types that appear in tables belonging to different topical domains (e.g. a hotel might have a phone number and an address, but not a release date and an artist). The pipeline uses two API calls: In the first step, we ask ChatGPT to predict the topical domain of the table to be annotated (e.g. music, hotels, restaurants, or events). In the second step, we include only the subset of all labels which are associated with the predicted domain to the task description and ask ChatGPT to annotate the columns of the table using only these semantic types. By breaking the label space in smaller spaces, we simplify the annotation task as the model is presented with less and more relevant labels to use for annotation. In the few-shot setup, in the first step we show tables and their domains as demonstrations, while in the second step we pick as demonstrations only tables from the predicted domain. An example of the two-step pipeline is shown in Figure \ref{fig:twostep}.

\textbf{Experimental Results.} The results for the two-step approach are summarized in Table \ref{tab:twostep}.  In all cases the table classification is an easy task and achieves an F1 score higher than 95\%. On average 1 error was made, and it involved a Hotel table that was predicted as an Event table. The hotel listed in this table contains the word ``Park" which seems to be a word that is also used in Events. 

The second step seems to achieve the highest performance in the zero-shot setup. As in the previous table format experiments (see Table \ref{tab:incontext}) where we didn't notice a performance increase with the increase of demonstrations, in this case as well we do not achieve higher performance when showing the model 4 demonstrations. The micro-F1 drops around 2\% in the 4-shot setup. This could be influenced by the quality/relatedness of table demonstrations to the test table as well as the length of the prompt.
Looking in detail at F1 scores for specific labels, ChatGPT predicts with 100\% F1 the types Duration, email, Country, currency, Coordinate, and Restaurant/name. Labels for which ChatGPT only reached F1 scores below 70\% include Photograph, Rating, LocationFeatureSpecification, and Time.


\begin{table}
  \caption{Results for the two-step approach in zero- and few-shot setups. ``S1-F1" refers to the average micro-F1 score reached by the first step and ``S2-F1" refers to the average micro-F1 score reached by the second step. The $\Delta$ F1 shows the difference between the micro-F1 score of our baseline model (column format zero-shot shown in the first line) to each experiment.}
  \label{tab:twostep}
  \begin{tabular}{cccccl}
    \toprule
    shots & S1-F1 & S2-P & S2-R & S2-F1&$\Delta$ F1\\
    \hline
    Baseline & - & 47.70 & 31.25 & 45.85& - \\
    \midrule
    0 & 95.56 & 90.08 & 86.60 & 89.47 & +43.62\\
    1 & 95.56 & 90.08 & 83.65 & 88.85 & +43.00 \\
    4 & 95.56 & 85.87 & 82.68 & 86.71 & +40.86 \\
  \bottomrule
\end{tabular}
\end{table}

\section{Comparison to Baselines}
\label{sec:baselines}

State of the art CTA methods~\cite{taSurvey2023} often rely on PLMs such as BERT~\cite{devlinBERTPretrainingDeep2019} and therefore require a significant amount of task-specific training examples. In this section, we compare the CTA results of ChatGPT to the results of different baseline methods with respect to training data efficiency.
We choose three baselines that cover different categories of machine learning methods: a Random Forest baseline, a fine-tuned RoBERTa model~\cite{liu_roberta_2019}, and DODUO~\cite{suhara2022annotating} a state of the art  method for CTA as well as column property annotation (CPA). DODUO fine-tunes BERT \cite{devlinBERTPretrainingDeep2019} using multi-task learning.

\textbf{Experimental Setup.} 
For the Random Forest baseline, we train the Random Forest using features generated with TF-IDF and we perform hyperparameter tuning using cross validation on the training set.
We fine-tune a RoBERTa \cite{liu_roberta_2019} model (\textit{roberta-base}) using the simple serialization method of concatenating all column values. We fine-tune for 30 epochs using a learning rate of 5e-5, a batch size of 32, and a maximum sequence length of 512.
Finally, we experiment with DODUO \cite{suhara2022annotating} and use its default parameters keeping the learning rate at 5e-5, training for 30 epochs and using a maximum length of 32 for the sequence. We change the default batch size from 16 to 32.

We experiment with different amounts of training examples for RoBERTa and DODUO, starting with a training set that contains 1 example per label (overall 32 examples) and going up to 50 training examples per label (overall 1600 examples). The training sets with 32, 159 and 1600 examples are also sampled from the original training set of the SOTAB CTA benchmark \cite{korini_sotab_2022}.

\begin{table}
  \caption{Baseline results using Random Forest, DODUO and RoBERTa models. ``shots" represents the number of demonstrations with which a model was trained on. The ChatGPT results in the first line correspond to the results of the zero-shot two-step approach and the $\Delta$ F1 shows the difference in micro-F1 scores between the ChatGPT model and the other models.}
  \label{tab:baselines}
  \begin{tabular}{lcccccc}
    \toprule
    Model&shots&P&R&F1& $\Delta$ F1\\ 
    \midrule
    ChatGPT & 0 & 90.08 & 86.60 & 89.47 & - \\\hline
    Forest & 159 & 38.36 & 43.75 & 46.15 & -43.32 \\ 
    Forest & 356 & 70.98 & 59.49  & 59.60 & -29.87\\ 
    RoBERTa& 32 &49.13 & 52.25&48.93 & -40.54 \\ 
    RoBERTa& 159 &82.41 & 81.79&79.2 & -10.27 \\ 
    RoBERTa& 356 & 90.87&87.70  &89.73 & +0.26 \\ 
    RoBERTa& 1600 & 87.59& 87.60 &86.79 & -2.68 \\ 
    DODUO& 356 & 1.95 & 48.92 & 6.37 & -83.10 \\ 
    DODUO& 1600 & 63.02 & 41.36  & 53.6 & -35.87 \\ 
  \bottomrule
\end{tabular}
\end{table}

\textbf{Baseline Results.} 
The results of the baseline experiments are shown in Table \ref{tab:baselines}. For DODUO and RoBERTa, we report the average of the results of three runs with different random seeds. The Random Forest baseline given 5 examples per label (159 examples in total) achieves a micro-F1 of 46.15\% and increasing the training examples the performance of the Random Forest increases. However, when given all the training set that contains 356 examples overall, the Random Forest still performs around 29\% less than the zero-shot version of ChatGPT. Fine-tuning a RoBERTa model on 32 training examples (one example per label) results in 48.93\% micro-F1, a comparable performance to the Random Forest with 159 examples, but around 40\% less than the zero-shot result of the table format of ChatGPT. When trained on 356 shots, the RoBERTa model increases to 89.73\% which surpasses the ChatGPT zero-shot setup \emph{table+inst+roles} (85.25\%, see Table \ref{tab:format}) and is comparable to the results of the zero-shot two-step pipeline (89.47\%, see Table \ref{tab:twostep}). Given more training examples, the performance decreases again probably due to the larger variety of tables that are contained in this larger set. Looking at the results for the state of the art method DODUO, we observe a lower performance when trained with 32 shots compared to the RoBERTa model. One of the reasons for this difference could be that RoBERTa is trained on text sequences which we also use for fine-tuning, so there is minimal difference between the input format of pre-training and fine-tuning, while for DODUO the serialization format changes to a table format which could require more training data to bridge the gap between pre-training and fine-tuning. When training DODUO with 1600 shots, we start seeing an increase in performance, but the difference to zero-shot ChatGPT setup still remains large at around 35\% less micro-F1.

From the results, we can conclude that ChatGPT in a zero-shot setup without any task-specific training examples is capable of reaching a CTA performance that is in the same range as the performance of PLM-based methods given hundreds of task-specific training examples, e.g. ChatGPT using the \emph{table+inst+roles} prompt reaches an F1 of 85.25\% (see Table \ref{tab:format}) while RoBERTa given 356 task-specific examples reaches an F1 of 89.73\% (+4.48\%). The difference is shortened to 0.26\% when the two-step pipeline in a zero-shot setting is used. If ChatGPT is provided with a single demonstration the difference shrinks to 1.29\%, e.g. the single-shot table prompt reaches an F1 of 88.44\% (see Table \ref{tab:incontext}).  
This conclusion is further underlined by the fact that ChatGPT only uses 5 rows of a table to reach a prediction, while RoBERTa uses on average 37 rows and DODUO uses on average 12 rows.

\section{Related Work}
\label{sec:rel-work}

This section gives an overview of related work on table annotation, data integration using large language models, as well as prompt engineering.

\textbf{Table Annotation and CTA.} Table annotation methods employ a wide range of different techniques ranging from statistical approaches to the more recent use of deep learning and pre-trained language models. Earlier deep learning methods like Sherlock \cite{sherlock-2019} and SATO \cite{sato-2020}, use column statistics and character distributions as features to their models. SATO is one of the first works to mention the importance of using not only intra-table context but also inter-table context by using topic vectors. TCN \cite{tcn-2021} uses a multi-task model trained on CTA and column relation prediction (CPA) that learns cell representations based on cells from the same table and cells from other tables. To predict the type of a column, the representations of the cells of one column are combined and passed through a classifier. RECA \cite{10.14778/3583140.3583149} continues the idea of using inter-table context by finding similar tables to the input table and uses the information in these related tables to find the correct type of a column. DODUO \cite{suhara2022annotating} is the state of the art method regarding CTA and CPA. The authors fine-tune a BERT \cite{devlinBERTPretrainingDeep2019} model using multi-task learning combining both tasks and introduce a new table serialization approach which passes a complete table to BERT by concatenating the content of all table columns. A further line of research focuses on learning table representations \cite{badaro:hal-03877085} and uses CTA as fine-tuning task for evaluating the learned representations: TURL \cite{deng_turl_2022} further pre-trains a TinyBERT \cite{jiao-etal-2020-tinybert} model on relational tables using Masked Language Modeling like BERT and Masked Entity Recovery as pre-training objectives to learn cell representations. The method is evaluated in 6 downstream tasks, one of which is CTA. TABBIE \cite{iida-etal-2021-tabbie} uses a transformer-based \cite{vaswani-transformer-2017} model to learn cell representations by using a cell corruption pre-training objective on both row and column level.

\textbf{Data Integration and LLMs.} LLMs have recently been employed to solve different tasks along the data integration pipeline. Jaimovitch-Lopez et al. \cite{jaimovitch-lopez_can_2022} explore the capability of LLMs to normalize different value formats such as dates and units of measurement and find that LLMs perform comparably with systems built specifically for this task. Tang et al.~\cite{tang2022generic} and Peeters and Bizer~\cite{Peeters2023ADBIS} explore using GPT-3 and ChatGPT for entity matching in zero- and few-shot setups. Narayan et al. \cite{foundationalWrangleVLDB2022}  conduct experiments with GPT-3 \cite{brown2020language} for entity matching, schema matching, data transformation, data imputation and error detection. They find that in a few-shot setting for all tasks, GPT-3 outperforms the state of the art methods, but they also argue that LLMs are sensible to differences in prompt formatting. Their schema matching experiments are the closest related work to the work presented in this paper. They experiment using the Synthea~\cite{zhang2021smat} dataset which includes schema data for tables and provides around 29,638 correspondences between them. They achieve an F1 score of 0.5\% in a zero-shot setting and 45.2\% in a three-shot setting. Unfortunately, their prompt design for the schema matching task isn't included in the paper so we can't compare it in detail to our own prompt designs.

\textbf{Prompt Engineering.} There are many works that experiment with prompt engineering~\cite{zhao2023survey, dong2023survey}.  Wei et al. \cite{wei_chain--thought_2022} show how decomposing a task into multiple subtasks and using them as demonstrations in \textit{chain-of-thought prompting} helps LLMs to achieve better results for reasoning tasks. Honovich et al. \cite{honovich2022instruction} experiment with using LLMs to generate instructions for various lexical and semantic tasks. In this work, we also experiment with human-written instructions with the difference that we choose to give the model multi-step instructions inspired by the chain-of-thought approach. Zhao et al. \cite{zhao2021calibrate} experiment with different prompt formats and discover that the choice of training examples and the choice of their ordering heavily influences the accuracy of the LLMs that they experiment with. They further try to understand what leads to these differences in accuracy and conclude that these models can be biased towards the majority class, towards the most recent training examples in the prompt, and towards tokens that appear frequently in the LLMs pre-training data. Closer to our work, in TabLLM \cite{pmlr-v206-hegselmann23a} the authors experiment with methods to serialize tabular data in prompts and compare their best method to deep learning and tree models. They find that representing columns using natural language sentences containing the feature name and feature value, e.g. `The price is 8.00 USD’, gave them the best performance in the zero-shot setup. Like these authors, we experiment with different approaches to present table columns to ChatGPT.

\section{Conclusion}
\label{sec:concl}

This paper is the first to apply large language models for the column type annotation task. We experiment with different prompt designs and compare the performance of ChatGPT to the performance of PLMs-based column type annotation methods.
We find ChatGPT to be much more training data efficient, requiring only a single demonstration to reach a similar performance as PMLs trained with hundreds of task-specific examples. In order to be able to deal with large label sets, we proposed decomposing the CTA task into a two-step pipeline consisting of first predicting the type of a table and second predicting column types using only the relevant subset of the full label set. 
As further work we plan to increase the difficulty of the test set by adding especially challenging tables from the SOTAB benchmark and to investigate how multi-step LLM pipelines as well as fine-tuned PLMs deal with these challenges.  

\bibliography{cta_chatgpt}




\end{document}